\documentclass{article} % For LaTeX2e
\usepackage{iclr2018_workshop,times}
\usepackage{hyperref}
\usepackage{url}

\usepackage{dsfont}
\usepackage{amsfonts}
\usepackage{amsmath}
\usepackage{slashbox}

\usepackage{caption}
\usepackage{subcaption}
\usepackage{tabularx,booktabs}
\usepackage{slashbox}
\usepackage{booktabs,caption}
\usepackage[flushleft]{threeparttable}
\usepackage{multirow}
\usepackage{mathtools}

\title{An interpretable LSTM neural network for autoregressive exogenous model}

\author{Tian Guo $^{*}$ \\
ETH Zurich, Switzerland \\
\texttt{\{tian.guo\}@gess.ethz.ch} \\
\And
Tao Lin
\thanks{Equal contribution.} 
\\
EPFL Lausanne, Switzerland \\
\texttt{\{tao.lin\}@epfl.ch} \\
\And
Yao Lu
\\
TalkingData Co., Ltd. Beijing, China \\
\texttt{\{yao.lu\}@tendcloud.com} \\
}

\begin{document}

\maketitle

\begin{abstract}
In this paper, we propose an interpretable LSTM recurrent neural network, i.e., multi-variable LSTM for time series with exogenous variables.  
Currently, widely used attention mechanism in recurrent neural networks mostly focuses on the temporal aspect of data and falls short of characterizing variable importance.
To this end, our multi-variable LSTM equipped with tensorized hidden states is developed to learn variable specific representations, which give rise to both temporal and variable level attention.
Preliminary experiments demonstrate comparable prediction performance of 
multi-variable LSTM w.r.t. encoder-decoder based baselines. More interestingly, variable importance in real datasets characterized by the variable attention is highly in line with that determined by statistical Granger causality test, which exhibits the prospect of multi-variable LSTM as a simple and uniform end-to-end framework for both forecasting and knowledge discovery. 
\end{abstract}

\section{Introduction}

Time series, a sequence of observations over time, is being generated in a wide variety of areas \citep{qin2017dual, lin2017hybrid, guo2018predicting}.
% Predicting future values of time series has been widely applied in many areas.
% e.g., financial market prediction [Wu et al., 2013], weather forecasting [Chakraborty et al., 2012], and complex dynamical system analysis [Liu and Hauskrecht, 2015]. 
% Recently, neural networks have been proven to be dramatical powerful in a wide spectrum of domains, e.g., natural language processing, computer vision, speech recognition, time series analysis, etc~\citep{wang2016morphological,sutskever2014sequence,yang2015deep,lipton2015learning,guo2016robust}. 
% A class of neural network architecture called Recurrent Neural Networks (RNN) is specifically designed to model sequential data. 
% Traditional RNNs suffer from the problem of vanishing gradients~\citep{bengio1994learning} and thus have difficulty in capturing long-term dependencies. 
% Among variants of RNN, 
Long short-term memory units (LSTM)~\citep{hochreiter1997long} and the gated recurrent unit (GRU)~\citep{cho2014properties} have achieved great success in various applications on sequence data because of the gate and memory mechanism \citep{wang2016morphological, lipton2015learning, guo2016robust}.

% The Recurrent layers with GRU [6] and LSTM [15] unit are carefully designed to memorize the historical information and hence to be aware of relatively long-term dependencies.
% they cannot model nonlinear relationships and do not differentiate among the exogenous (driving) input terms.
% In many settings, data is collected as multiple time series, where each recorded time series is an observation of some underlying dynamical process of interest. These observations are often time-marked with known event times, and one desires to do a range of standard analyse
% the values of the target series by using autoregressive part itself and exogenous series

In this paper, we focus on time series with exogenous variables. Specifically, given a target time series, we have an additional set of time series corresponding to exogenous variables. A predictive model using the historical data of both target and exogenous variables to predict the future values of the target variable is an autoregressive exogenous model, referred to as ARX. In addition to forecasting, it is also highly desirable to distill knowledge via the model, e.g., understanding the different importance of exogenous variables w.r.t. the evolution of the target series \citep{hu2018listening, siggiridou2016granger, zhou2015probabilistic}.
% The success of RNN motivates us to consider using RNN, LSTM in our case, to realize both accurate forecasting of the target series and importance interpretation of exogenous variables w.r.t the target one. 
However, current LSTM RNN falls short of such capability. 
When fed with the historical observations of the target and exogenous variables, LSTM blindly blends the information of all variables into the hidden states and memory cells for subsequent prediction. Therefore, it is intractable to distinguish the contribution of individual variables by looking into hidden states \citep{zhang2017stock}. 

Recently, attention-based neural networks~\citep{bahdanau2014neural,xu2015show,chorowski2015attention} have been proposed to enhance the ability of LSTM in using long-term memory as well as the interpretability. The attention mechanism is mostly applied to hidden states across time steps, thereby solely uncovering the temporal level importance rather than the variable level importance. 

To this end, we propose an interpretable LSTM recurrent neural network, called multi-variable LSTM, for ARX problem. A distinguishing feature of our multi-variable LSTM is to enable each neuron of the recurrent layer to encode information exclusively from a certain variable. As a result,
from the overall hidden states of the recurrent layer, we derive variable specific hidden representations over time steps, which can be flexibly used for forecasting and temporal-variable level attentions.

% \citep{vaswani2017attention}

% and use such specific hidden representations flexibly. Multi-variable LSTM provides a simple and uniform end-to-end framework to achieve forecasting and knowledge discovery simultaneously with one recurrent layer. 

% It is very flexible to use such input specific hidden representations to achieve   

% Moreover, we develop variate level attention mechanism on such hidden state subsets to select relevant exogenous series to make predictions as well as uncovering the variable importance. 

% The Nonlinear autoregressive exogenous (NARX)
% model, which predicts the current value of a time
% series based upon its previous values as well as the

% ATTENTION

% But, the gain in accuracy from use of RNNs is at the cost of model output that is notoriously difficult to interpret. 

% The key idea of the attention mechanism is to perform prediction by using weighted sum of hidden states across all the time steps. The weights dependent on the hidden states provide explicit interpretation about the importance of previous steps w.r.t. the predicted step. 
% introduce extra source of information to guide the extraction of sentence
% Recently, a hierarchical attention network [Yang et al., 2016], which uses two layers of attention mechanism to select relevant encoder hidden states across all the time steps, was also developed.
% self-attention mechanism allows extracting different aspects

\section{Related work}
The success of attention mechanism proposed in~\citep{bahdanau2014neural} has motivated a wide use of attention in image processing~\citep{ba2014multiple,mnih2014recurrent,gregor2015draw,xu2015show}, natural language processing~\citep{hermann2015teaching,rush2015neural,lin2017structured} and speech recognition~\citep{chorowski2015attention}.
However, traditional attention mechanism is normally applied to hidden states across time steps, thereby failing to reveal variable level attention.  
% especially for sequences with heterogeneous natures.
Only some very recent studies~\citep{choi2016retain,qin2017dual} attempted to develop attention mechanism to handle multi-variable sequence data. 
Both of them build on top of encoder-decoder architecture, and make the prediction using pre-weighted input obtained from the encoder or an additional RNN. Our MV-LSTM is a simple one recurrent layer architecture enabling variable specific representations. 

% \citep{choi2016retain} develops two-level neural attention model for time series classification, trying to capture the influential time step and significant features.
% \citep{qin2017dual} proposes dual-stage attention-based recurrent neural network,
% where input attention mechanism on the encoder adaptively builds input features for each time step based on previous hidden state, while temporal attention mechanism selects relevant encoder hidden states across all time steps.

% None of the previous studies, to the best of our knowledge, clearly explored how to interpret spatial-temporal importance of the time series via the attention model.

% This is precisely the focus of the present study, that introduces generalizations of the content based attention model to capture pseudo-periods and improve forecasting in time series.

\section{Multi-variable LSTM}
In this section, we present the proposed multi-variable LSTM referred to as MV-LSTM in detail. 
% \tao{should we include a figure for multi-variable LSTM? we can consider to shorten the introduction.}

Assume we have $N-1$ exogenous time series and target series $\mathbf{y}$ of length $T$, where $\mathbf{y} = ( y_1, \cdots, y_T )$ and $\mathbf{y} \in \mathbb{R}^T$.\footnote{
Vectors are assumed to be in column form throughout this paper.
} 
By stacking exogenous time series and target series, we define the multi-variable input of MV-LSTM at each time step as
$\mathbf{X} = ( \mathbf{x}_1, \ldots, \mathbf{x}_T )$, 
where $ \mathbf{x}_t = ( x_{t,1}, \ldots, x_{t, N-1}, y_{t} )  \in \mathbb{R}^N$ and $x_{t,n} \in \mathbb{R}$ is the observation of $n$-th exogenous time series at time $t$.
% each dimension corresponds to the observations of a variable. since each hidden state ingest information from all input variable, has to develop seperated rnn to individual model 
Given $\mathbf{X}$, we aim to learn a non-linear mapping to the one-step ahead value of the target series, namely $\hat{y}_{T+1} = \mathcal{F}(\mathbf{X})$,
where $\mathcal{F}(\cdot)$ represents the MV-LSTM neural network we present below. 

% kaiser2017depthwise, kuchaiev2017factorization, 
Inspired by \citep{he2017wider}, our MV-LSTM has tensorized hidden states and the update scheme ensures that each element of the hidden state tensor encapsulates information exclusively from a certain variable of the input.
% , while it is updated by exploiting multi-variable input together.

Specifically, we define the hidden state and memory cell for $t$-th time step of MV-LSTM as 
$\mathbf{h}_t \in \mathbb{R}^M$ and $\mathbf{c}_t \in \mathbb{R}^M$, 
where $M$ is the size of recurrent layer. 
$\mathbf{h}_t$ is tensorized as 
$\mathcal{H}_t = [ \mathbf{h}_t^{1}, \ldots, \mathbf{h}_t^{N} ]^\top $, 
where $\mathcal{H}_t \in \mathbb{R}^{N \times d}$, $\mathbf{h}_t^{n} \in \mathbb{R}^d$ and $N \cdot d = M$. 
The element $\mathbf{h}_t^{n}$ of tensor $\mathcal{H}_t$ is a variable specific representation corresponding to $n$-th input dimension.
% as well as evolving correlatively. 
% The update process in MV-LSTM presented below ensures that such representations
% as the gates for updating are derived by using all input dimensions.
We further define the input-to-hidden transition tensor as 
$\mathcal{W}_x = [ \mathbf{W}_x^{1}, \ldots, \mathbf{W}_x^{N} ]^\top $, 
where $\mathcal{W}_x \in \mathbb{R}^{N \times d }$ and $\mathbf{W}_x^{n} \in \mathbb{R}^d $. 
The hidden-to-hidden transition tensor is defined as: 
$\mathcal{W}_h = [ \mathbf{W}_h^{1}, \ldots, \mathbf{W}_h^{N} ]^\top$, 
where $\mathcal{W}_h \in \mathbb{R}^{N \times d \times d}$ and 
$\mathbf{W}_h^{n} \in \mathbb{R}^{d \times d}$.

% characterised by its ability to have different recurrent transition functions for each possible input, w

Given the new incoming input $\mathbf{x}_t$ and the hidden state $\mathbf{h}_{t-1}$ up to $t-1$,
% and $\mathbf{c}_{t-1}$,
we formulate the iterative update process by using $\mathcal{W}_x$ and $\mathcal{W}_h$ as:
\begin{align*}
   \mathbf{j}_t = \tanh( \mathcal{H}_{t - 1} * \mathcal{W}_h + \mathbf{x}_t * \mathcal{W}_x + \mathbf{b}_j )&= \tanh \left( \begin{bmatrix}
           (\mathbf{W}_h^1 \mathbf{h}_{t-1}^1)^\top \\
           \vdots \\
            (\mathbf{W}_h^N \mathbf{h}_{t-1}^N)^\top 
         \end{bmatrix} 
         +
         \begin{bmatrix}
           (\mathbf{W}_x^1 x_{t, 1})^\top \\
           \vdots \\
           (\mathbf{W}_x^N x_{t, N})^\top
         \end{bmatrix} 
         + \mathbf{b}_j \right)
\end{align*} 
where $*$ represents the element-wise multiplication operation on tensor elements. 
$\mathcal{H}_{t-1} * \mathcal{W}_h \in \mathbb{R}^{N \times d}$ 
is the concatenation of $N$ product results of hidden tensor element $\mathbf{h}_t^{n}$
and the corresponding transition matrix $\mathbf{W}_h^{n}$. 
Likewise, 
% the result of 
$ \mathbf{x}_t * \mathcal{W}_x \in \mathbb{R}^{N \times d}$,
% is a $M$ dimensional vector, 
and represents how the current input $\mathbf{x}_t$ update the hidden state. 

$\mathbf{j}_t$ is an $N \times d$ dimensional tensor, and each $d$-dimensional element corresponds to one input variable, 
% as well as encoding information exclusively from that variable. 
encoding the information exclusively from the variable specific hidden state and the input variable.
% Therefore, elements in $\mathbf{j}_t$ can be grouped the same way as tensorized $\mathcal{H}_t$, each of which corresponds to a certain variable. 
% Different from the tensorized update of $\mathbf{j}_t$,

The input, forget and output gates in MV-LSTM are updated by using all input dimensions of $\mathbf{x}_t$, so as to utilize the cross-correlation between multi-variable time series. In particular, $[\mathbf{i}_t, \mathbf{f}_t, \mathbf{o}_t]^{\top} = \sigma \left( \mathbf{W}  [\mathbf{x}_t, \mathbf{h}_{t-1}] + \mathbf{b} \right)$.
% \begin{align}
% \begin{bmatrix}
% \mathbf{i}_t \\
% \mathbf{f}_t \\
% \mathbf{o}_t \\
% \end{bmatrix}
% &= \sigma \left( \mathbf{W}  [\mathbf{x}_t, \mathbf{h}_{t-1}] + \mathbf{b} \right) 
% \end{align} 
The updated memory cell and hidden states are obtained from $\mathbf{c}_t = \mathbf{f}_t \odot \mathbf{c}_t + \mathbf{i}_t \odot \mathbf{\tilde{j}}_t $, where $\mathbf{\tilde{j}}_t \in \mathbb{R}^M $ is the flattened vector of $\mathbf{j}_t$. Then, $\mathbf{h}_t = \mathbf{o}_t \odot \tanh(\mathbf{c}_t)$.

After feeding $\mathbf{x}_T$ into MV-LSTM, 
we obtain the hidden representation $\mathbf{h}_T^n$ w.r.t. each variable, which can be combined with the attention mechanism to predict $y_{T+1}$ 
as well as interpreting variable importance. Concretely, the attention process is as: $e^n = \tanh( \mathbf{W}_e \mathbf{h}_{T}^n + b_e )$ and $\alpha^n = \frac{exp(e^n)}{\sum_{k=1}^N exp(e^k)}$. Then, the prediction is derived as: $\hat{y}_{T+1} = \sum_{n=1}^N \alpha^n ( \mathbf{W_n} \mathbf{h}_n^{\top} + b_n )$.
Note that MV-LSTM is able to apply temporal attention with ease. In the present work, we focus on evaluating variable attention. 
% \tao{should we add more words to attention part?}

% Wider and Deeper, Cheaper and Faster: Tensorized LSTMs for Sequence Learning
% Tensorizing Neural Networks
% FACTORIZATION TRICKS FOR LSTM NETWORKS
\section{Experiments}
In this part, we report some preliminary results to demonstrate the prediction performance of MV-LSTM as well as the variable importance interpretation. Please refer to the appendix section for more results about MAE errors and variable attention interpretation. 

We use two real datasets. 
\textbf{PM2.5:}
It contains hourly PM2.5 data and the associated meteorological data in Beijing of China. PM2.5 measurement is the target series. The exogenous time series include dew point, temperature, pressure, combined wind direction, cumulated wind speed, hours of snow, and hours of rain. 
% \textbf{PLANT:}
% This dataset records the time series of energy production of a photovoltaic (PV) power plant in Italy. Exogenous data consists of $19$ dimensional time series regarding weather conditions (such as temperature, irradiance, cloud coverage, etc.) over the geographical area of the plant. 
% Each time point is the hourly aggregation obtained as the average of all the measures available in a specified hour.
% Training data spans over a temporal period of 12 months (year 2012) including the daily target time series (power observed for each plant), whereas testing data consists of 3 months (January to March 2013) for which the target time series (power) is not provided.
\textbf{ENERGY:} It collects the appliance energy use in a low energy building. The target series is the energy data logged every 10 minutes. Exogenous time series consist of $14$ variables, e.g., house inside temperature conditions and outside weather information including temperature, wind speed, humanity and dew point from the nearest weather station.

% , gradient boosted tree(GBT) \citep{friedman2001greedy}, 
Baselines include ensemble methods on time series, i.e., random forests (RF) \citep{liaw2002classification, meek2002autoregressive} and extreme gradient boosting (XGT) \citep{chen2016xgboost, friedman2001greedy}, and state-of-the-art attention based recurrent neural networks on multi-variable sequence data, referred to as DUAL \citep{qin2017dual} and RETAIN \citep{choi2016retain}.
% \tian{encoder decoder}

In the first group of experiments, we report the prediction performance of all approaches in Table~\ref{tab:horizon_rmse}. 
% \backslashbox{Dataset}{Model}
\begin{table}[!h]
  \centering
  \caption{Test errors (RMSE)}
  \begin{tabular}{|c|c|c|c|c|c|}
    \hline
    Dataset & RF & XGT & DUAL & RETAIN & MV-LSTM \\
    \hline
%     \hline
    PM2.5&   $0.573 \pm 0.003$& $0.370 \pm 0.002$& $0.355 \pm 0.002$& $1.112 \pm 0.017$ &$0.340 \pm 0.001$\\
    ENERGY& $0.494 \pm 0.004$& $0.360 \pm 0.003$& $0.372 \pm 0.006$& $0.669 \pm 0.006$& $0.361 \pm 0.001$\\
    \hline
\end{tabular}
\label{tab:horizon_rmse}
\end{table}

\begin{figure*}[!hbt]
\centering
\includegraphics[width=0.8\textwidth]{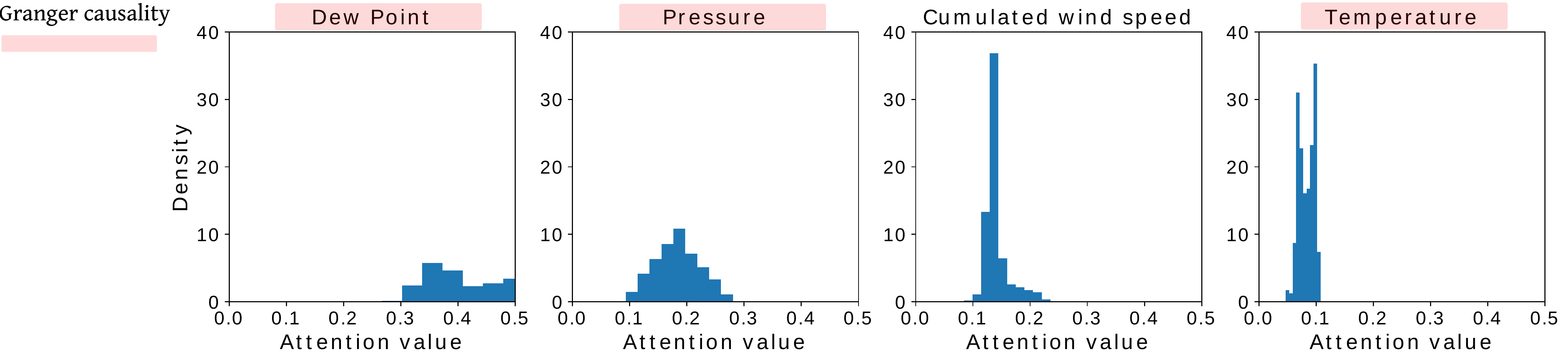}
\caption{Top four variables ranked by empirical mean of attention values in MV-LSTM on PM2.5 dataset. Variable names with colored background indicates the identification by Granger-causality as well.
% that the variable is identified by Granger-causality test as well.
}
\label{fig:pm}
\end{figure*}

Next, we analyze the variable level attention obtained in MV-LSTM on PM2.5 dataset. 
Specifically, in testing phase MV-LSTM outputs variable attention values specific for each testing instance, and thus for each variable we can estimate an empirical distribution of the corresponding attention value. 
Figure~\ref{fig:pm} shows the histograms of the attention values on top four variables, which are ranked by the empirical mean of their attention values.
% Specifically, for each testing data instance fed into MV-LSTM, we obtain the associated variable level attention and thus we use the empirical attention distribution of data instances corresponding to each variable to rank variables, as is shown in Fig. \ref{fig:pm}. 
For comparison, variables identified by Granger causality test \citep{arnold2007temporal} w.r.t. the target variable are shown with a colored background. 

We observe in Figure~\ref{fig:pm} that three variables (i.e., Dew point, pressure, and temperature) identified by Granger causality test are also top ranked by the variable attention of MV-LSTM. 
As pointed out by~\citep{liang2015assessing}, dew point and pressure are usually affected by the arrival of the northerly wind, which brings in drier and fresher air. This is exactly in line with the observation in the rank of such three variables by MV-LSTM. On the contrary, DUAL fails to reveal meaningful variable importance, as is shown in the appendix section. 

% The pressure distribution in July was much lower than that in the other three months. The meteorological variables are mutually correlated. 
 
% Indeed, a decrease in the dew point and an increase in the pressure are usually accompanied by the arrival of the northerly wind, which brings in drier and fresher air. We need a systematical approach to model the overall effect of the weather variables on PM2.5, which we deal with in the next section.

\bibliography{iclr2018_workshop}

\begin{thebibliography}{29}
\providecommand{\natexlab}[1]{#1}
\providecommand{\url}[1]{\texttt{#1}}
\expandafter\ifx\csname urlstyle\endcsname\relax
  \providecommand{\doi}[1]{doi: #1}\else
  \providecommand{\doi}{doi: \begingroup \urlstyle{rm}\Url}\fi

\bibitem[Arnold et~al.(2007)Arnold, Liu, and Abe]{arnold2007temporal}
Andrew Arnold, Yan Liu, and Naoki Abe.
\newblock Temporal causal modeling with graphical granger methods.
\newblock In \emph{Proceedings of the 13th ACM SIGKDD international conference
  on Knowledge discovery and data mining}, pp.\  66--75. ACM, 2007.

\bibitem[Ba et~al.(2014)Ba, Mnih, and Kavukcuoglu]{ba2014multiple}
Jimmy Ba, Volodymyr Mnih, and Koray Kavukcuoglu.
\newblock Multiple object recognition with visual attention.
\newblock \emph{arXiv preprint arXiv:1412.7755}, 2014.

\bibitem[Bahdanau et~al.(2014)Bahdanau, Cho, and Bengio]{bahdanau2014neural}
Dzmitry Bahdanau, Kyunghyun Cho, and Yoshua Bengio.
\newblock Neural machine translation by jointly learning to align and
  translate.
\newblock \emph{arXiv preprint arXiv:1409.0473}, 2014.

\bibitem[Chen \& Guestrin(2016)Chen and Guestrin]{chen2016xgboost}
Tianqi Chen and Carlos Guestrin.
\newblock Xgboost: A scalable tree boosting system.
\newblock In \emph{SIGKDD}, pp.\  785--794. ACM, 2016.

\bibitem[Cho et~al.(2014)Cho, Van~Merri{\"e}nboer, Bahdanau, and
  Bengio]{cho2014properties}
Kyunghyun Cho, Bart Van~Merri{\"e}nboer, Dzmitry Bahdanau, and Yoshua Bengio.
\newblock On the properties of neural machine translation: Encoder-decoder
  approaches.
\newblock \emph{arXiv preprint arXiv:1409.1259}, 2014.

\bibitem[Choi et~al.(2016)Choi, Bahadori, Sun, Kulas, Schuetz, and
  Stewart]{choi2016retain}
Edward Choi, Mohammad~Taha Bahadori, Jimeng Sun, Joshua Kulas, Andy Schuetz,
  and Walter Stewart.
\newblock Retain: An interpretable predictive model for healthcare using
  reverse time attention mechanism.
\newblock In \emph{Advances in Neural Information Processing Systems}, pp.\
  3504--3512, 2016.

\bibitem[Chorowski et~al.(2015)Chorowski, Bahdanau, Serdyuk, Cho, and
  Bengio]{chorowski2015attention}
Jan~K Chorowski, Dzmitry Bahdanau, Dmitriy Serdyuk, Kyunghyun Cho, and Yoshua
  Bengio.
\newblock Attention-based models for speech recognition.
\newblock In \emph{Advances in neural information processing systems}, pp.\
  577--585, 2015.

\bibitem[Friedman(2001)]{friedman2001greedy}
Jerome~H Friedman.
\newblock Greedy function approximation: a gradient boosting machine.
\newblock \emph{Annals of statistics}, pp.\  1189--1232, 2001.

\bibitem[Gregor et~al.(2015)Gregor, Danihelka, Graves, Rezende, and
  Wierstra]{gregor2015draw}
Karol Gregor, Ivo Danihelka, Alex Graves, Danilo~Jimenez Rezende, and Daan
  Wierstra.
\newblock Draw: A recurrent neural network for image generation.
\newblock \emph{arXiv preprint arXiv:1502.04623}, 2015.

\bibitem[Guo \& Antulov-Fantulin(2018)Guo and
  Antulov-Fantulin]{guo2018predicting}
Tian Guo and Nino Antulov-Fantulin.
\newblock Predicting short-term bitcoin price fluctuations from buy and sell
  orders.
\newblock \emph{arXiv preprint arXiv:1802.04065}, 2018.

\bibitem[Guo et~al.(2016)Guo, Xu, Yao, Chen, Aberer, and Funaya]{guo2016robust}
Tian Guo, Zhao Xu, Xin Yao, Haifeng Chen, Karl Aberer, and Koichi Funaya.
\newblock Robust online time series prediction with recurrent neural networks.
\newblock In \emph{2016 IEEE DSAA}, pp.\  816--825. IEEE, 2016.

\bibitem[He et~al.(2017)He, Gao, Xiao, Liu, He, and Barber]{he2017wider}
Zhen He, Shaobing Gao, Liang Xiao, Daxue Liu, Hangen He, and David Barber.
\newblock Wider and deeper, cheaper and faster: Tensorized lstms for sequence
  learning.
\newblock In \emph{Advances in Neural Information Processing Systems}, pp.\
  1--11, 2017.

\bibitem[Hermann et~al.(2015)Hermann, Kocisky, Grefenstette, Espeholt, Kay,
  Suleyman, and Blunsom]{hermann2015teaching}
Karl~Moritz Hermann, Tomas Kocisky, Edward Grefenstette, Lasse Espeholt, Will
  Kay, Mustafa Suleyman, and Phil Blunsom.
\newblock Teaching machines to read and comprehend.
\newblock In \emph{Advances in Neural Information Processing Systems}, pp.\
  1693--1701, 2015.

\bibitem[Hochreiter \& Schmidhuber(1997)Hochreiter and
  Schmidhuber]{hochreiter1997long}
Sepp Hochreiter and J{\"u}rgen Schmidhuber.
\newblock Long short-term memory.
\newblock \emph{Neural computation}, 9\penalty0 (8):\penalty0 1735--1780, 1997.

\bibitem[Hu et~al.(2018)Hu, Liu, Bian, Liu, and Liu]{hu2018listening}
Ziniu Hu, Weiqing Liu, Jiang Bian, Xuanzhe Liu, and Tie-Yan Liu.
\newblock Listening to chaotic whispers: A deep learning framework for
  news-oriented stock trend prediction.
\newblock In \emph{Proceedings of the Eleventh ACM International Conference on
  Web Search and Data Mining}, pp.\  261--269. ACM, 2018.

\bibitem[Liang et~al.(2015)Liang, Zou, Guo, Li, Zhang, Zhang, Huang, and
  Chen]{liang2015assessing}
Xuan Liang, Tao Zou, Bin Guo, Shuo Li, Haozhe Zhang, Shuyi Zhang, Hui Huang,
  and Song~Xi Chen.
\newblock Assessing beijing's pm2. 5 pollution: severity, weather impact, apec
  and winter heating.
\newblock In \emph{Proc. R. Soc. A}, volume 471, pp.\  20150257. The Royal
  Society, 2015.

\bibitem[Liaw et~al.(2002)Liaw, Wiener, et~al.]{liaw2002classification}
Andy Liaw, Matthew Wiener, et~al.
\newblock Classification and regression by randomforest.
\newblock \emph{R news}, 2\penalty0 (3):\penalty0 18--22, 2002.

\bibitem[Lin et~al.(2017{\natexlab{a}})Lin, Guo, and Aberer]{lin2017hybrid}
Tao Lin, Tian Guo, and Karl Aberer.
\newblock Hybrid neural networks for learning the trend in time series.
\newblock In \emph{Proceedings of the Twenty-Sixth International Joint
  Conference on Artificial Intelligence, IJCAI-17}, pp.\  2273--2279,
  2017{\natexlab{a}}.

\bibitem[Lin et~al.(2017{\natexlab{b}})Lin, Feng, Santos, Yu, Xiang, Zhou, and
  Bengio]{lin2017structured}
Zhouhan Lin, Minwei Feng, Cicero Nogueira~dos Santos, Mo~Yu, Bing Xiang, Bowen
  Zhou, and Yoshua Bengio.
\newblock A structured self-attentive sentence embedding.
\newblock \emph{arXiv preprint arXiv:1703.03130}, 2017{\natexlab{b}}.

\bibitem[Lipton et~al.(2015)Lipton, Kale, Elkan, and
  Wetzell]{lipton2015learning}
Zachary~C Lipton, David~C Kale, Charles Elkan, and Randall Wetzell.
\newblock Learning to diagnose with lstm recurrent neural networks.
\newblock \emph{arXiv preprint arXiv:1511.03677}, 2015.

\bibitem[Meek et~al.(2002)Meek, Chickering, and
  Heckerman]{meek2002autoregressive}
Christopher Meek, David~Maxwell Chickering, and David Heckerman.
\newblock Autoregressive tree models for time-series analysis.
\newblock In \emph{SDM}, pp.\  229--244. SIAM, 2002.

\bibitem[Mnih et~al.(2014)Mnih, Heess, Graves, et~al.]{mnih2014recurrent}
Volodymyr Mnih, Nicolas Heess, Alex Graves, et~al.
\newblock Recurrent models of visual attention.
\newblock In \emph{Advances in neural information processing systems}, pp.\
  2204--2212, 2014.

\bibitem[Qin et~al.(2017)Qin, Song, Cheng, Cheng, Jiang, and
  Cottrell]{qin2017dual}
Yao Qin, Dongjin Song, Haifeng Cheng, Wei Cheng, Guofei Jiang, and Garrison~W.
  Cottrell.
\newblock A dual-stage attention-based recurrent neural network for time series
  prediction.
\newblock In \emph{Proceedings of the 26th International Joint Conference on
  Artificial Intelligence}, IJCAI'17, pp.\  2627--2633. AAAI Press, 2017.

\bibitem[Rush et~al.(2015)Rush, Chopra, and Weston]{rush2015neural}
Alexander~M Rush, Sumit Chopra, and Jason Weston.
\newblock A neural attention model for abstractive sentence summarization.
\newblock \emph{arXiv preprint arXiv:1509.00685}, 2015.

\bibitem[Siggiridou \& Kugiumtzis(2016)Siggiridou and
  Kugiumtzis]{siggiridou2016granger}
Elsa Siggiridou and Dimitris Kugiumtzis.
\newblock Granger causality in multivariate time series using a time-ordered
  restricted vector autoregressive model.
\newblock \emph{IEEE Transactions on Signal Processing}, 64\penalty0
  (7):\penalty0 1759--1773, 2016.

\bibitem[Wang et~al.(2016)Wang, Cao, Xia, and de~Melo]{wang2016morphological}
Linlin Wang, Zhu Cao, Yu~Xia, and Gerard de~Melo.
\newblock Morphological segmentation with window lstm neural networks.
\newblock In \emph{AAAI}, 2016.

\bibitem[Xu et~al.(2015)Xu, Ba, Kiros, Cho, Courville, Salakhudinov, Zemel, and
  Bengio]{xu2015show}
Kelvin Xu, Jimmy Ba, Ryan Kiros, Kyunghyun Cho, Aaron Courville, Ruslan
  Salakhudinov, Rich Zemel, and Yoshua Bengio.
\newblock Show, attend and tell: Neural image caption generation with visual
  attention.
\newblock In \emph{International Conference on Machine Learning}, pp.\
  2048--2057, 2015.

\bibitem[Zhang et~al.(2017)Zhang, Aggarwal, and Qi]{zhang2017stock}
Liheng Zhang, Charu Aggarwal, and Guo-Jun Qi.
\newblock Stock price prediction via discovering multi-frequency trading
  patterns.
\newblock In \emph{Proceedings of the 23rd ACM SIGKDD International Conference
  on Knowledge Discovery and Data Mining}, pp.\  2141--2149. ACM, 2017.

\bibitem[Zhou et~al.(2015)Zhou, Huang, Zhang, Hu, Du, Song, and
  Xie]{zhou2015probabilistic}
Xiabing Zhou, Wenhao Huang, Ni~Zhang, Weisong Hu, Sizhen Du, Guojie Song, and
  Kunqing Xie.
\newblock Probabilistic dynamic causal model for temporal data.
\newblock In \emph{Neural Networks (IJCNN), 2015 International Joint Conference
  on}, pp.\  1--8. IEEE, 2015.

\end{thebibliography}
\bibliographystyle{iclr2018_workshop}

\newpage
\section{Appendix}

\begin{table}[!hbt]
  \centering
  \caption{Test errors (MAE)}
  \begin{tabular}{|c|c|c|c|c|c|}
    \hline
    Dataset & RF & XGT & DUAL & RETAIN & MV-LSTM \\
    \hline
%     \hline
    PM2.5&   $0.433 \pm 0.011$& $0.302 \pm 0.012$& $0.248 \pm 0.003$& $0.943 \pm 0.018$ &$0.227 \pm 0.002$\\
    ENERGY& $0.404 \pm 0.021$& $0.310 \pm 0.023$& $0.249 \pm 0.006$& $0.507 \pm 0.016$& $0.256 \pm 0.006$\\
    \hline
\end{tabular}
\label{tab:horizon_mae}
\end{table}

\begin{figure*}[!hbt]
\centering
\includegraphics[width=0.8\textwidth]{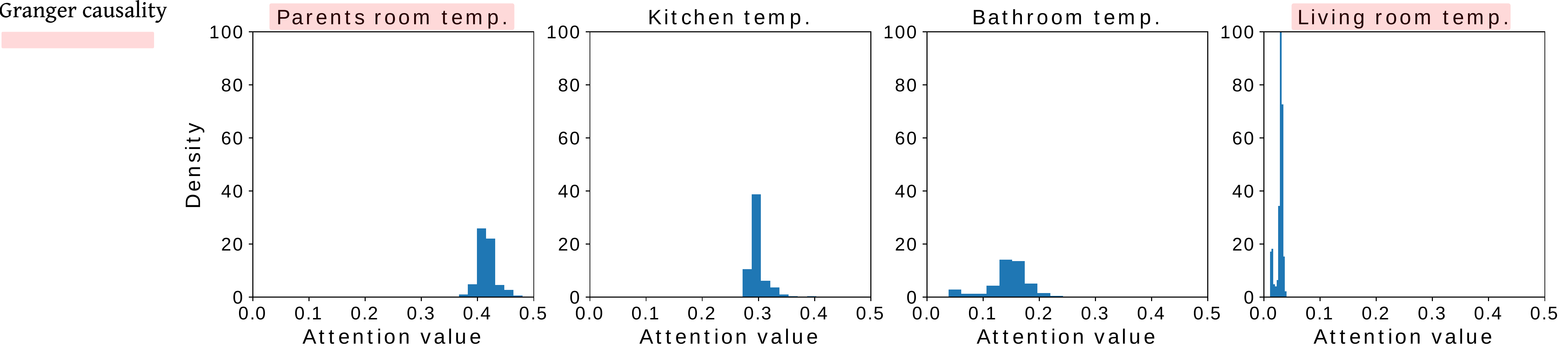}
\caption{Top four variables ranked by empirical mean of attention values in MV-LSTM on ENERGY dataset. Variable name with the color background indicate that the variable is identified by Granger-causality test as well. Only variable parents room temp. and living room temp. are identified by Granger-causality test in this dataset. }
\label{fig:energy_mv}
\end{figure*}
\begin{figure*}[!hbt]
\centering
\includegraphics[width=0.8\textwidth]{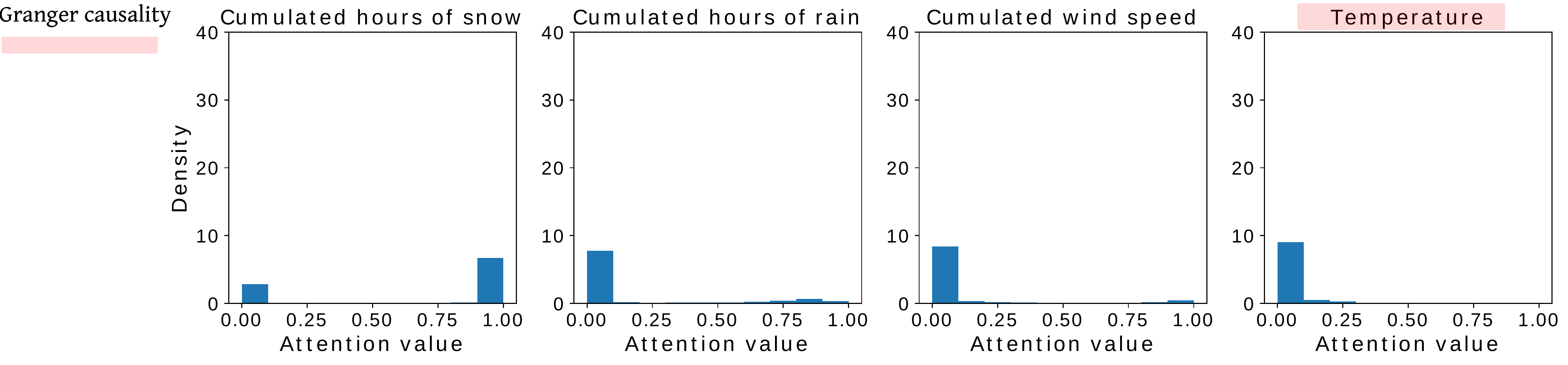}
\caption{Top four variables ranked by empirical mean of attention values in DUAL on PM2.5 dataset. Variable name with the color background indicate that the variable is identified by Granger-causality test as well.}
\label{fig:pm_dual}
\end{figure*}
\begin{figure*}[!hbt]
\centering
\includegraphics[width=0.8\textwidth]{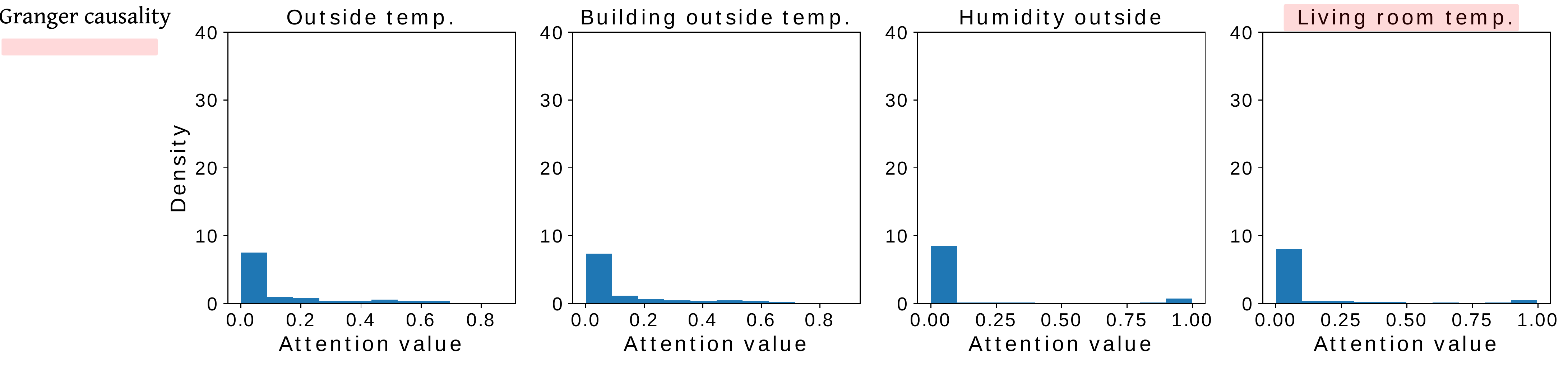}
\caption{Top four variables ranked by empirical mean of attention values in DUAL on ENERGY dataset. Variable name with the color background indicate that the variable is identified by Granger-causality test as well.}
\label{fig:energy_dual}
\end{figure*}

\end{document}